\documentclass{article}

    \usepackage[preprint]{neurips_2023}

\usepackage[utf8]{inputenc} 
\usepackage[T1]{fontenc}    
\usepackage{hyperref}       
\usepackage{url}            
\usepackage{booktabs}       
\usepackage{amsfonts}       
\usepackage{nicefrac}       
\usepackage{microtype}      
\usepackage{xcolor}         
\usepackage{graphicx}
\usepackage{wrapfig}
\usepackage{amsmath}

\title{Robust Multi-Modal Forecasting: Integrating Static and Dynamic Features}

\author{%
  Jeremy Qin\\
  Department of Computer Science and Operational Research\\
  Université de Montréal - Mila\\
  \texttt{jeremy.qin@umontreal.ca} \\
}

\begin{document}
\maketitle

\begin{abstract}
Time series forecasting plays a crucial role in various applications, particularly in healthcare, where accurate predictions of future health trajectories can significantly impact clinical decision-making. Ensuring transparency and explainability of the models responsible for these tasks is essential for their adoption in critical settings. Recent work has explored a top-down approach to bi-level transparency, focusing on understanding trends and properties of predicted time series using static features. In this work, we extend this framework by incorporating exogenous time series features alongside static features in a structured manner, while maintaining cohesive interpretation. Our approach leverages the insights of trajectory comprehension to introduce an encoding mechanism for exogenous time series, where they are decomposed into meaningful trends and properties, enabling the extraction of interpretable patterns. Through experiments on several synthetic datasets, we demonstrate that our approach remains predictive while preserving interpretability and robustness. This work represents a step towards developing robust, and generalized time series forecasting models. The code is available at \url{https://github.com/jeremy-qin/TIMEVIEW}
\end{abstract}
\section{Introduction}
\label{sec-intro}
With the development of high-performance models, their benefits have driven adoption across various domains. However, the translational value of machine learning models has not been as widely realized in high-stakes areas such as healthcare. This is primarily due to concerns about the trustworthiness and transparency of these models, stemming from their complex architecture and black-box nature.  Recently, many works have begun to explore top-down approaches to AI transparency. \citep{zou2023representationengineeringtopdownapproach} characterized the emerging field of representation engineering, a top-down approach that emphasizes population-level representations as the main component of analysis, enabling the explanation and control of high-level concepts in deep neural networks. Additionally, \citep{Kacprzyk.TransparentTimeSeries.2024} proposed a bi-level transparency framework for time series forecasting to facilitate understanding of predicted trajectories. In fact, in the essay of Nobel Laureate P.W. Anderson "More Is Different", he describes how complex phenomena cannot be explained simply from the bottom-up, and that examining from the top-down allows to uncover generalizable rules. These developments suggest a promising new direction for AI transparency research.

Time series model allow for different types of inputs, such as static features, time series and exogenous features. While previous works mainly focused on static features for better interpretability and formalism, we focus on integrating dynamic features such as time series while maintaining transparency and interpretability of the predicted trajectories. This multi-modality is particularly relevant in domains like healthcare, where features such as blood pressure, glucose levels, and sleep activity, measured over time, are crucial for interpreting the illness trajectory. For instance, \citep{niu2023deep} highlight the importance of combining time series data with clinical records to improve mortality prediction in intensive care units. However, understanding how changes in dynamic inputs affect the model’s predictions is not as straightforward. For static features, it is natural to ask, "What would happen to the model's prediction if this feature changed?" Yet, it is less intuitive to ask similar questions for dynamic features, such as, "What would happen to the model's prediction if this feature had a different value two weeks ago?" This challenge underscores the need for a transparent way to integrate dynamic features, allowing for more intuitive and interpretable analysis of their impact on the model’s output.

Furthermore, we address a critical concern of robustness and fairness when using time series features to model future trajectories in domains such as healthcare and physics. This issue arises from the inherent measurement uncertainty associated with the instruments and situation used to record various metrics. For instance, \citep{Jackson2023} provides evidence that vital sign measurement are subject to value preference to a degree that it could potentially affect conclusions. Similarly \citep{Mari2024} examined the effect of measurement uncertainty on fault diagnosis systems and found that accounting for this uncertainty in neural networks improved the reliability of the models. In healthcare, this random error in measurement is referred as precision, which is formally defined by the International Vocabulary of Metrology as "closeness of agreement between indications or measured quantity value obtained by replicated measurements on the same of similar objects under specified conditions". A practical example of this is that a model's predictions should remain consistent even if a patient's blood pressure is measured at different hospitals or using different instruments. This illustrates how measurement uncertainty can raise significant concerns about a model's robustness and fairness. To address this, we propose leveraging insights on trajectory comprehension from \citep{Kacprzyk.TransparentTimeSeries.2024} to encode time series features into trends and properties. This approach enhances both robustness and transparency by focusing on the underlying patterns rather than the variability inherent in individual measurements.

In summary, our contributions are:
\begin{enumerate}
    \item We extend existing framework TIMEVIEW by incorporating dynamic time series data alongside static features.
    \item We propose an encoding mechanism that decomposes dynamic features into trends and properties, which allows for the extraction of interpretable patterns. This helps address measurement uncertainty by focusing on robust trend analysis rather than individual data points, thus enhancing model reliability.
    \item We leverage contrastive learning to align latent representations of static and dynamic features, ensuring consistent and robust model behavior.
\end{enumerate}

\section{Time Series Forecasting}
\label{sec-background}
Time series forecasting has gained significant attention due to its broad applicability across various domains, including healthcare, physics, and finance. While many studies have focused separately on either time series features or static features, real-world applications often rely on multiple modalities. This is especially true in healthcare, where a patient’s profile is built over different time intervals or cycles. For example, static features like gender and age provide basic information, while dynamic measurements—such as daily glucose levels, blood pressure, or heart rates recorded between consultations—offer insights into the patient’s ongoing condition. Observing the trends in these dynamic features can reveal whether a treatment is effective or if a patient’s health is improving. Therefore, to generalize effectively to real-life scenarios, it is essential to consider settings that integrate both static and dynamic features, enabling a more comprehensive and realistic modeling approach.

\subsection{Problem formulation}
We adopt the problem formulation given by \citep{Kacprzyk.TransparentTimeSeries.2024} to include both static and dynamic features. Let $T \in \mathbb{R}$ represent the time horizon. Each sample consists of static features $x_s^{(n)} \in \mathbb{R}^M$, where $M \in \mathbb{N}$ is the number of static features, and a set of dynamic features $x_d^{(n)} \in \mathbb{R}^{t'^{(n)} \times  D}$, where $t'^{(n)} \in \mathbb{R}^{D}$ is the number of past time steps for the n-th sample and $D \in \mathbb{N}$ is the number of dynamic features. The samples are associated with discrete trajectories $y^{(n)} \in \mathbb{R}^{N_n}$ observed at times $t^{(n)} \in \mathbb{R}^{N_n}$ where $N_n \in \mathbb{N}$ is the number of measurements for the n-th sample. For simplicity, we assume that $t'(n)$ is constant across all samples and dynamic features. Then, given a dataset $\{x_s^{(n)}, x_d^{(n)}, y^{(n)}, t^{(n)}\}$, our aim is to model the relationship between these static and dynamic inputs and the resulting trajectory.

\section{Method}
\label{sec-method}
Based on the problem formulation outlined in Section \ref{sec-background} and the formalism introduced by \citep{Kacprzyk.TransparentTimeSeries.2024}, we propose a method to integrate time series features in a transparent and interpretable manner. Additionally, we introduce the use of contrastive learning to effectively align the representations of both static and dynamic modalities, ensuring consistency and enhancing the model’s robustness. Our work builds on top of the TIMEVIEW framework presented in \citep{Kacprzyk.TransparentTimeSeries.2024}.
\subsection{TIMEVIEW}
TIMEVIEW is a framework that combines a predictive model using B-Spline basis functions with an algorithm for creating composition maps, enhancing interpretability of time series predictions. It achieves bi-level transparency by linking feature vectors to both the compositions of predicted trajectories and their transition points, using cubic splines to represent these trajectories. Each trajectory is expressed as a linear combination of B-Splines, defined by a latent vector, allowing easy calculation of compositions through cubic polynomials associated with intervals determined by internal knots. This approach facilitates understanding of the overall trajectory by extracting and visualizing key dynamical motifs.

\subsection{Interpretable Dynamic Features Encoding}
Integrating time series features as an additional modality can be approached in various ways. To preserve the interpretability of the TIMEVIEW framework, we adopt a dual encoding mechanism: one encoder for static features and a recurrent encoder for dynamic features, designed to capture temporal relationships. However, directly encoding the raw data points of dynamic features does not facilitate interpretable reasoning about the output trajectory, nor does it support intuitive counterfactual analysis.

\textbf{Real-world Settings}: In real-world settings, it is neither practical nor intuitive to pose "what if" or "how to" questions about individual changes in time series inputs. Instead, it is often more meaningful to explore scenarios where trends or properties over time have changed.  For example, asking, What if glucose levels showed an increasing trend after we adjusted the treatment dosage during the previous cycle?" is a more realistic concern than, "What if glucose levels changed two weeks ago?". Motivated by this distinction, we make the assumption that small changes in time series features values should not impact future trajectories unless there is a change in underlying trends and properties. Here we refer to trends refer to the shape or motifs of the curve over specific intervals, while properties denote the transition points between different intervals, such as maximum and minimum values or other significant metrics.

This assumption is also motivated by the inherent measurement uncertainty present in data acquisition in real-life scenarios especially in domains when there is different standards and instruments. In the healthcare domain, it is common that different hospitals have different instruments or methods to measure patient's vital signs which necessarily will engender discrepancies in measurements. However, we postulate that even if the measurements might be different depending on the instruments, place or situation of the patient, the trends and properties should stay the same. As such, we propose an interpretable dynamic features encoding method to first encode the different time series features into trends and properties so that only a change in these will give different representations.

\textbf{Integrating Trends and Properties Through Interleaving}: To further enhance the structure of our dynamic encoding, we ensure that trends and their corresponding properties maintain a direct correlation. Trends are determined by analyzing the first and second derivatives of the curve within each interval, capturing the overall direction and changes in the data. Properties are derived from the transition points between intervals, such as peaks, troughs, and inflection points, which highlight significant metrics or changes. We leave the formalism of trends and properties to \citep{Kacprzyk.TransparentTimeSeries.2024}, which provides a detailed explanation of their definitions. Then, instead of treating all trends and properties separately, we interleave them during the encoding process to preserve their natural association. For example, if we have trends $I_1, I_2, ..., I_8$ and properties $P_1, P_2, ..., P_8$, we generate an interleaved vector: $[I_1, P_1, I_2, P_2, ..., I_8, P_8]$. This approach ensures that each trend remains closely linked with its corresponding property, which is critical for maintaining interpretability and temporal correlation.


\textbf{Architecture}: To adapt our method to the TIMEVIEW framework, we still use the same static encoder $\textbf{h}: \mathbb{R}^M \rightarrow \mathbb{R}^N$ to match the latent vector $\textbf{c} \in \mathbb{R}^B$ that describes each spline, and we add a recurrent encoder $\textbf{r} : \mathbb{R}^{t \times D} \rightarrow \mathbb{R}^B$. To combine these two latent vectors, we have $m = h + h_{dynamic}$, and we define our model $g : (\mathbb{R}^M, \mathbb{R}^{t \times D}) \rightarrow \mathbf{\hat{Y}}$ as
$$g(x_s, x_d)(t) =  \hat{y}_{x_s, x_d}(t) = \sum_{b=1}^B m(x_s, x_d)_b \phi_b(t)$$

For the static encoder, we use the same implementation as TIMEVIEW, and for the dynamic encoder we use a recurrent neural network. By having this architecture, we make sure to be able to leverage the transparency and interpretability of the TIMEVIEW model as it does not impact the composition extraction.

\subsection{Contrastive Loss Regularization}
Introducing an additional modality allows us to enrich the representations of each sample. By assuming that the static and dynamic latent vectors provide a consistent representation of the same sample; for example, in a healthcare setting, both static features (e.g., age, gender) and dynamic features (e.g., blood pressure readings over time) should cohesively reflect the overall health status of a patient, we use contrastive learning to align these representations. This is achieved by encouraging similar latent vectors to be closer in the representation space, while pushing dissimilar vectors apart. The contrastive loss is computed using a cross-entropy loss over the similarity matrix $\textbf{s}$:
\begin{align*}
    L_c&= - \frac{1}{N} \sum_{i=1}^N \log \frac{\exp(s_{ii})}{\sum_{j=1}^N \exp(s_{ij})} \\
    &= - \frac{1}{N} \sum_{i=1}^N \log \frac{\exp \left( \frac{\bar{h}^{(i)}_s \bar{h}^{(i)}_d}{\tau} \right)}{\sum_{j=1}^N \exp \left( \frac{\bar{h}^{(i)}_s \bar{h}^{(j)}_d}{\tau} \right)}
\end{align*}
$\bar{h}_s$ and $\bar{h}_d$ are the normalized static and dynamic latent vectors, and $\tau$ is a temperature parameter that controls the scale of the similarity scores. We incorporate this contrastive loss into the overall objective function, which consists of a Mean Squared Error (MSE) loss and a L2 regularization term . The combined objective function is:
$$\mathbf{L} = \frac{1}{N} \sum_{n=1}^N \left(\frac{1}{N_n} \sum_{j=1}^{N_n} \left(y_j^d - \sum_{b=1}^B m(x_s, x_d)_b \phi_b(t^d_j) \right)^2 \right) + \alpha L_{L2}(g) + \beta L_c$$
The inclusion of the contrastive loss not only aligns the static and dynamic representations but also regularizes the model by constraining how the compositions change when the input features are varied. This encourages a consistent and interpretable behavior across both modalities.

\subsection{Formalization of Robustness}
Ensuring robustness in dynamic feature encoding is crucial for reliable predictions, especially in domains like healthcare, where data variability and measurement noise are common. Our approach formalizes robustness by encoding dynamic features into trends and properties that are resistant to small perturbations, ensuring that the model’s predictions remain stable unless there is a significant change in the underlying trends or properties. In the following sections, we use sensitivity and robustness as complementing concepts.

\textbf{Robustness Criterion for Raw Time Series Features} \\
Let $x_d(t)$ represent a raw dynamic feature over time, and let $\hat{y}$ be the model's output prediction. When using raw time series features, any small perturbation $\epsilon(t)$ can directly impact $x_d(t)$:
$$x_d(t) = x_{d, true}(t) + \epsilon(t)$$
The sensitivity of the model output to these perturbations can be expressed as:
$$\left| \frac{\partial\hat{y}}{\partial \epsilon(t)}\right| \neq 0$$
where $\hat{y}$ is sensitive to every small change $\epsilon(t)$ because these small perturbations can still propagate through the dynamic encoder.

\textbf{Formalism of Robustness :} \\
When the raw time series features are processed directly by the dynamic encoder, in this case let's assume we use a LSTM:
$$h_{raw} = LSTM([x_d(t_1), x_d(t_2), ..., x_d(t_n)])$$
each perturbation $\epsilon(t)$ in $x_d(t)$ can propagate through the LSTM. Thus, we have:
$$\frac{\partial h_{raw}}{\partial \epsilon(t)} \neq 0$$
Now, when we consider encoding a time series feature by constructing a combined representation by interleaving trends and properties, we can write it as:
$$Z = [I_1, P_1, I_2, P_2, ..., I_N, P_N]$$
where we divided the time series $x_d(t)$ into N intervals and each $I_i$ represents a segment of the time series. By doing so, trends $I_i$ aggregate the behaviour over intervals, reducing the effect of small local perturbations such that:
$$\frac{\partial T_i}{\partial \epsilon(t)} \approx 0, \ \forall t \in I_i$$
The trend encodes broader patterns over each interval, ensuring that the dynamic encoder can focus on smoothed behaviors, leading to more robust processing. Furthermore, we have that properties $P_i$ capture key transitions, inherently focusing on changes that are significant over time. Hence, we have that:
$$\frac{\partial P_i}{\partial \epsilon(t)} \approx 0$$
Then, we can write the sensitivity of the encoded interleaved feature to small perturbations as:
\begin{align*}
    \frac{\partial h_{dynamic}}{\partial \epsilon(t)} &= \frac{\partial h_{dynamic}}{\partial I_i} \cdot \frac{\partial I_i}{\partial \epsilon(t)} + \frac{\partial h_{dynamic}}{\partial P_i} \cdot \frac{\partial P_i}{\partial \epsilon(t)} \approx 0
\end{align*}

\textbf{Quantifying Sensitivity}\\
We define sensitivity of raw time series features as:
$$R_{raw} = \mathbb{E}_t \left[ \left| \frac{\partial \hat{y}}{\partial \epsilon(t)}\right| \right]_{raw} >  0$$
where $\hat{y}$ are the predicted trajectories that depends on $h_{raw}$. 
For trend-property encoding, we define sensitivity as:
$$R_{trend} = \mathbb{E}_t \left[ \left| \frac{\partial \hat{y}}{\partial \epsilon(t)}\right| \right]_{trend}$$
Given that:
$$\frac{\partial \hat{y}}{\partial \epsilon(t)} = \frac{\partial \hat{y}}{\partial h_{dynamic}} \cdot \frac{\partial h_{dynamic}}{\partial \epsilon(t)}$$
and knowing that $\frac{\partial h_{dynamic}}{\partial \epsilon(t)} \approx 0$, we finally have that:
$$R_{trend} \ll R_{raw}$$
which indicates that trend-property encoding reduces sensitivity, leading to more robust representations. Note that to arrive at this, we make the assumption that the predicted trajectories are more sensitive with regards to the raw time series latent vector $h_{raw}$ as opposed to the trend-property encoded one $h_{trend}$ because $h_{raw}$ directly reflects the time-dependent variability of $x_d(t)$.

\section{Related Works}
\label{sec-relworks}
\subsection{Transparent and Explainable Time-Series Models}
Recent advancements in time series explainability have focused on enhancing transparency by developing methods that provide clearer insights into how models derive predictions. \citep{liu2024timexlearningtimeseriesexplanations} introduced TIMEX++, which addresses issues like distributional shifts through a modified information bottleneck approach, ensuring that explanations remain consistent and interpretable. TIMEX, proposed by \citep{queen2023encodingtimeseriesexplanationsselfsupervised}, employs a consistency-based surrogate model to maintain faithful representations in the latent space, offering discrete and interpretable attribution maps. \citep{xu2024kolmogorovarnoldnetworkstimeseries} developed Kolmogorov-Arnold Networks (KAN), which use spline-parametrized functions to improve interpretability by dynamically capturing relationships, with variants designed to explain concept drift and multivariate patterns. \citep{schlegel2021tsmulelocalinterpretablemodelagnostic} extended the LIME framework with specialized segmentation techniques, enhancing the local transparency of explanations for complex time series models. Lastly, the Generalized Additive Time Series Model (GATSM) by \citep{kim2024transparentnetworksmultivariatetime} combines feature networks and transparent temporal modules, providing a clear and interpretable structure that matches the performance of more opaque, black-box models.

\subsection{Multi-modal Time Series}
Real-life healthcare scenarios often involve multiple data modalities, including static features like demographics, clinical text, imaging, and continuous vital signs, making it essential to develop methods that can effectively integrate these diverse sources. \citep{niu2023deep} combines Bio-BERT embeddings of clinical text with LSTM-encoded time-series data, alongside static patient information, using a fusion module to capture cross-modal correlations for improved mortality prediction in ICUs. MultiSurv by \citep{vale2021long} addresses pan-cancer survival prediction by integrating clinical, imaging, and omics data through dedicated submodels and a fusion layer, seamlessly handling static and dynamic inputs, even with missing data. Additionally, \citep{KRONES2025102690} also discuss the importance of efficient data fusion techniques to combine imaging, text, time series, and tabular data at different stages.
\section{Experiments and Results}
\label{sec-empirical}
To evaluate the proposed method for integrating time series features with static features in the TIMEVIEW model, and to understand the contribution of each component, we conducted a series of experiments comparing its performance against models using raw time series features. As such, we focus solely on the TIMEVIEW model and test the addition of each component. The experiments were designed to demonstrate the benefits of smoothing temporal patterns and incorporating transition properties, as well as leveraging this additional modality to align representations between static and dynamic features. We report results across multiple synthetic datasets and present ablation studies to analyze the impact of each component.

\textbf{Datasets} Experiments were conducted on 3 synthetic datasets (D-Sine, D-Beta and D-Tumor) based on those presented in \citep{Kacprzyk.TransparentTimeSeries.2024}, which originally included only static features. For D-Sine and D-Beta, we crafted single time series features of constant length for simplicity, while D-Tumor was designed to incorporate slightly more controlled temporal patterns. Specifically, D-Tumor included three distinct time series features representing past blood pressure levels, glucose levels, and oxygen saturation. To closely replicate real-life settings, we controlled the mean and variance of these features to reflect realistic ranges of values. The ground truth for the tumor volume trajectories was calculated by adapting the model from \citep{wilkerson2017estimation}, adjusting it with factors derived from the time series features. Since we use synthetic datasets where we controlled the ground truth to be dependent on the time series features, we do not compare our methods to simply TIMEVIEW with static features, as it is not fair comparison.

\begin{table}[h!]
\centering
\caption{Comparison of TIMEVIEW model variants across three synthetic datasets: D-Sine, D-Beta, and D-Tumor. Each row represents a different model configuration, from raw time series features to versions with encoded trends, transition properties, contrastive learning (CL), and positional encoding (PE). We use mean squared error as the metric for comparison. Each configuration was ran on 5 different seeds and we report the mean and standard deviation.}
\label{table:performance}
\begin{tabular}{lccc}
\toprule
\multicolumn{1}{c}{Methods}                          & \multicolumn{1}{c}{D-Sine} & \multicolumn{1}{c}{D-Beta} & \multicolumn{1}{c}{D-Tumor} \\ \midrule
TIMEVIEW + Raw Time Series                           & $0.173 \pm  0.057$           & $0.093 \pm  0.015$                      & $0.082 \pm  0.021$                       \\
TIMEVIEW + Trends                            & $0.171 \pm  0.034$           & $0.076 \pm  0.003$                      & $0.087 \pm  0.017$                       \\
TIMEVIEW + Trends and Properties             & $0.151 \pm  0.035$           & $0.078 \pm  0.003$                      & $0.082 \pm  0.025$                       \\
TIMEVIEW + Trends and Properties + CL        & $0.163 \pm  0.042$           & $0.082 \pm  0.006$                      & $0.077 \pm  0.015$                       \\
\bottomrule
\end{tabular}
\end{table}

\textbf{Ablation Studies} To understand the contribution of each component in the TIMEVIEW model, we conducted ablation studies. Starting with a baseline version that utilized raw time series features, we progressively added encoded trends, transition properties, contrastive learning (CL), and positional embeddings (PE). This step-by-step approach allowed us to isolate and evaluate the effect of each component on model performance. The results, as shown in Table \ref{table:performance}, indicate that adding encoded trends generally improved performance across the datasets, and incorporating transition properties further enhanced it. While the addition of contrastive learning had mixed effects, suggesting that its impact may vary depending on the dataset characteristics, we believe this is due to the simplicity of the synthetic datasets used. Although the improvements are not consistently observed across all configurations, the overall trend shows that the proposed methods lead to better performance compared to the baseline using raw time series features. 
 
\section{Conclusion}
\label{sec-concl}
In this work, we proposed a method for integrating time series features with static features within the TIMEVIEW model, aiming to enhance robustness, interpretability, and alignment. By encoding dynamic features as trends and transition properties, our approach captures meaningful temporal patterns while also accounting for measurement uncertainty, making the model more robust to variability in the data. The interleaved representation ensures compatibility with TIMEVIEW's existing architecture, preserving transparency and interpretability. This integration maintains the model’s ability to provide clear insights into the decision-making process, enabling intuitive and reliable analysis.

Through our experiments on synthetic datasets, we demonstrated that this method leads to overall improved performance compared to relying solely on raw time series inputs. Additionally, the use of trends and properties allowed dynamic features to remain interpretable, enabling intuitive counterfactual reasoning. This means that the model can simulate and explain important questions on changes in dynamic inputs, such as past health indicators, would affect the predicted outcomes, offering valuable insights for real-world applications.



 \bibliographystyle{named}
\bibliography{ref}







\end{document}